%% file: main.tex
\pdfoutput=1

\documentclass[11pt]{article}

\usepackage{acl}

\usepackage{times}
\usepackage{latexsym}
\usepackage{graphicx}
\usepackage[T1]{fontenc}

\usepackage[utf8]{inputenc}
\usepackage{mathtools}
\usepackage{microtype}
\usepackage{enumitem}
\usepackage{multirow}
\usepackage{comment}
\usepackage{booktabs} 
\usepackage{array}
\usepackage{arydshln}
\newcolumntype{L}[1]{>{\raggedright\let\newline\\\arraybackslash\hspace{0pt}}m{#1}}
\newcolumntype{C}[1]{>{\centering\let\newline\\\arraybackslash\hspace{0pt}}m{#1}}
\newcolumntype{R}[1]{>{\raggedleft\let\newline\\\arraybackslash\hspace{0pt}}m{#1}}

\title{On the Evaluation of Answer-Agnostic Paragraph-level Multi-Question Generation}

\author{Jishnu Ray Chowdhury \\
  University of Illinois at Chicago \\
  \texttt{jraych2@uic.edu} \\\And
  Debanjan Mahata \\
  Moody's Analytics \\
  \texttt{debanjan.mahata@moodys.com} \\
  \AND 
  {\bf Cornelia Caragea} \\
  University of Illinois at Chicago \\
  \texttt{cornelia@uic.edu}}

\begin{document}
\maketitle
\begin{abstract}
We study the task of predicting a set of salient questions from a given paragraph without any prior knowledge of the precise answer. We make two main contributions. First, we propose a new method to evaluate a set of predicted questions against the set of references by using the Hungarian algorithm to assign predicted questions to references before scoring the assigned pairs. We show that our proposed evaluation strategy has better theoretical and practical properties compared to prior methods because it can properly account for the coverage of references. Second, we compare different strategies to utilize a pre-trained seq2seq model to generate and select a set of questions related to a given paragraph. The code is available\footnote{\url{https://github.com/JRC1995/QuestionGenerationPub}}. 

\end{abstract}

\input{introduction}

\input{task_definition}
\input{evaluation}

\input{method}

\input{rank_and_select}
\input{experiments}

\input{related_work}
\input{conclusion}

\bibliography{anthology,custom}
\bibliographystyle{acl_natbib}

\end{document}

%% file: introduction.tex
\section{Introduction}
\label{sec:intro}
Question generation (QG) is the task of automatically generating questions that are pertinent to a given text. QG has multiple applications. For example, it can be used to construct educational materials by generating practice questions to test reading comprehension \cite{heilman2010good}, help dialogue systems ask better questions \cite{wang2018learning}, or drive conversations around news articles \cite{laban2020whats}. QG has been also used for creating synthetic data to train competitive question answering models \cite{shakeri2020end, puri2020training}, for pre-training models to improve downstream performance \cite{narayan2020qurious}, and for evaluating the factuality of automated abstractive summarizations \cite{wang2020asking}. For any large-scale application on emerging data, it would be infeasible to manually construct questions; thus, the need for automating QG arises. There are, however, multiple variants of QG. In this paper, we focus on a particular variant of the task (see Figure \ref{fig:example}) with the following characteristics:
\setlength{\parskip}{1.0em}

\begin{figure}[t]

\begin{small}
\fbox{\parbox{0.45\textwidth}{
\textbf{Input:} {the central government estimates that over 7,000 inadequately engineered schoolrooms collapsed in the earthquake chinese citizens have since invented a catch phrase tofu-dregs schoolhouses.....}

\vspace{1mm}

\textbf{Generated Questions:}
\begin{enumerate}[nolistsep]
    \item what policy caused many families to lose their only child?
    \item what is the catch phrase for inadequately engineered schoolhouses?
    \item how many inadequately engineered schoolrooms collapsed in the earthquake?
    \item what is the age of the so-called illegal children?
\end{enumerate}

\vspace{1mm}

\textbf{Reference Questions:}
\begin{enumerate}[nolistsep]
\item how many schoolrooms collapsed in the quake?
\item what catch-phrase was invented as a result of collapsed schools?
\item why did so many schools collapse during the earthquake?
\item what are the estimations of how many schoolrooms collapsed?
\item what has the citizenry started calling these type of schools?
\item what can illegal children be registered as in place of their dead siblings?
\end{enumerate}
}}
\end{small}
\caption{A sample of a passage from SQuAD1.1 with a set of generated questions and a set of references}
\vspace{-5mm}
\label{fig:example}
\end{figure}

\noindent \textbf{1. Answer-Agnostic} - Given a document, our models are trained to generate questions related to the document without any prior knowledge about the answer (answer-agnostic QG). The model has to learn to (implicitly or explicitly) seek out question-worthy sentences and phrases to generate their corresponding questions. Answer-agnostic QG is, in general, more challenging than answer-aware question generation (in which the answer is usually highlighted in some fashion). Practical applications of answer-aware QG can be partly limited because they would require some external mechanism (for example a named entity tagger \cite{wang2020asking} or a keyphrase generator \cite{willis2019keyphrase}) or manual annotation to choose specific answers to generate questions from. An external mechanism may be only able to pick very specific types of answers (like keyphrases or named entities) and thus restrict the range of questions that can be generated. 

\noindent \textbf{2. Paragraph-Level} - Instead of generating questions at a sentence-level, we focus on generating the most important questions for a whole paragraph (or a document). The model has to learn to decide which areas on the paragraph it should focus while generating the most salient questions. 


\noindent \textbf{3. Multiple Questions} - 
We focus on generating and evaluating a \textit{set} of \textit{multiple} questions that can be asked from the given paragraph. 
\setlength{\parskip}{1.0em}

\noindent Overall, similar to \cite{du2017identifying}, our aim with this task is to generate an appropriate number of diverse questions around the most important (``question-worthy") areas in the paragraph. In Figure \ref{fig:example}, we show an example that reflects the above characteristics of the task. As can be seen in the figure, we have a paragraph input, a set of ground references, and a set of generated questions, ideally corresponding to the salient question-worthy areas of the paragraph. However, although we aim to generate a \textit{set} of questions comparable to a ground truth \textit{set}, standard evaluations of QG \cite{du2017learning,qi2020prophetnet, lopez2021transformerqg} based on BLEU \cite{papineni2002bleu}, METEOR \cite{banerjee-lavie-2005-meteor}, ROUGE-L \cite{lin-2004-rouge}, or Q-BLEU \cite{nema2018towards} are only designed to measure the quality of a \textit{single} generated question against a set of references per sample. These metrics are not designed to compare a \textit{set} of generated questions against the set of references (see $\S$\ref{sec:multi_evaluation}). 
For proper evaluation, we propose a novel evaluation scheme which utilizes the Hungarian algorithm \cite{kuhn1955hungarian} to first optimally assign each generated question to a particular ground truth question before comparing the assigned pairs. 
\setlength{\parskip}{0em}

Besides evaluation, as another contribution, we provide a systematic empirical comparison (under a common framework) of multiple task-specific strategies based on generation granularity-level ($\S$\ref{sec:generation_granularity}) and multiplicity (or mode) ($\S$\ref{sec:generation_mode}).  
Given that state of the art is generally achieved by pre-trained Transformer-based models \cite{varanasi2020copybert,qi2020prophetnet,lopez2021transformerqg}, we investigate whether direct brute use of pre-trained models is sufficient or some of the prior strategies are still relevant to improve overall performance.

%% file: task_definition.tex
\section{Task Definition}
\label{sec:task_def}
For our task, as input we have a paragraph $P$ as a sequence of sentences $P = (s_i)_{i \in [1,n]}$ where each sentence $s_i$ is a sequence of word (or sub-word) tokens $\left (s_i = (t_j)_{j \in [1,m]}\right )$. The desired output is a set of questions $Q = \{q_l | l \in [1,p]\}$ where each question $q_i$ is a sequence of word (or sub-word) tokens $\left (q_l = (t_j)_{j \in [1,r]}\right )$. We assume that each question $q_l$ have its answer information in the given paragraph $P$. We also assume that for any question $q_l$, $P$ contains at least one sentence $s_i$ that have sufficient information to answer the question $q_l$. Although, a sensible question (for example, questions asking for relevant missing information) can have its answer absent from the paragraph, in this work, we solely focus on generating questions that are answerable from the information in the input. 

Throughout this paper, we assume that the provided set of ground truth questions contains the ideal questions to generate and also, the ideal \textit{number} of questions to generate. The assumption may be violated in practice because, depending on the annotation process, the given ground truth set may not reflect the ideals; moreover, in certain applications the ideal number of questions to be generated may need to be defined by the user or some specific rules. However, given the difficulty of accounting for all these variables, we stick to the aforementioned assumption. 

%% file: evaluation.tex
\section{Multi-Question Evaluation}
\label{sec:multi_evaluation}
Typically, standard text generation evaluation $n$-gram match metrics like BLEU, ROUGE, and METEOR are used for QG. \citet{nema2018towards} proposed interpolating $n$-gram match metrics with an answerability score creating Q-BLEU/Q-METEOR/Q-ROUGE as more suitable metrics for evaluating questions. All these $n$-gram match metrics are still designed to evaluate a single generated sequence against a set of ground truth reference sequences per sample. However, we aim to compare a \textit{set} of multiple generated questions against the set of references per sample (e.g., as shown in Figure \ref{fig:example}). A simple approach to do this is to calculate an $n$-gram match score (BLEU, METEOR, etc.) individually for each prediction in the set and then average the results. However, such average-metrics have at least two major limitations:
\setlength{\parskip}{1.0em}

\begin{figure}[t]
\centering
\begin{small}
\fbox{\parbox{0.45\textwidth}{
\textbf{Case 1: Number mismatch}

\textbf{Generated Questions (Set 1):}
\begin{enumerate}[nolistsep]
    \item who is the current president of the united states?
\end{enumerate}

\vspace{1mm}
\textbf{Case 2: Reference miscoverage}

\textbf{Generated Questions (Set 2):}
\begin{enumerate}[nolistsep]
    \item who is the current president of the united states?
    \item who is the president of the united states now?
    \item who is the president of the united states currently?
\end{enumerate}

\vspace{1mm}

\textbf{Reference Questions:}
\begin{enumerate}[nolistsep]
 \item who is the current president of the united states?
 \item when was the great wall of china built?
 \item how does the business model of wikipedia work? 
\end{enumerate}
}}
\end{small}
\caption{Example cases where a high average n-gram match score can be achieved despite the set of predictions being significantly different from the set of references. In case 1, there are too few predictions. In case 2, the generated predictions are all paraphrases and none are close to reference 2 or 3.}
\vspace{-5mm}
\label{fig:error_case_examples}
\end{figure}

\begin{figure*}[t]
\centering
\begin{small}
\fbox{\parbox{\dimexpr\textwidth-2\fboxsep-2\fboxrule}{

\textbf{Prediction:} what policy caused many families to lose their only child? 
   
$\xrightarrow{\text{assigned to}}$ \textbf{Reference:} why did so many schools collapse during the earthquake? (\textbf{METEOR:} $9.33$)

\mbox{}\hspace{-\fboxsep}%
\makebox[\dimexpr\linewidth+2\fboxsep\relax]{\hrulefill}
    
\textbf{Prediction:} what is the catch phrase for inadequately engineered schoolhouses? 

$\xrightarrow{\text{assigned to}}$ \textbf{Reference:} what catch-phrase was invented as a result of collapsed schools? (\textbf{METEOR:} $18.19$)

\mbox{}\hspace{-\fboxsep}%
\makebox[\dimexpr\linewidth+2\fboxsep\relax]{\hrulefill}

\textbf{Prediction:} how many inadequately engineered schoolrooms collapsed in the earthquake? 

$\xrightarrow{\text{assigned to}}$ \textbf{Reference:} how many schoolrooms collapsed in the quake? (\textbf{METEOR:} $48.83$)

\mbox{}\hspace{-\fboxsep}%
\makebox[\dimexpr\linewidth+2\fboxsep\relax]{\hrulefill}

\textbf{Prediction:} what is the age of the so-called illegal children?

$\xrightarrow{\text{assigned to}}$ \textbf{Reference:} what can illegal children be registered as in place of their dead siblings? (\textbf{METEOR:} $16.46$)

\mbox{}\hspace{-\fboxsep}%
\makebox[\dimexpr\linewidth+2\fboxsep\relax]{\hrulefill}

\textbf{Average METEOR:} $23.20$; \textbf{Overall Match Score (S)}: $92.81$;  \textbf{Multi-METEOR:} $18.56$ 
}}
\end{small}
\caption{Example optimal assignment based on METEOR using the Hungarian algorithm when the set of predictions (generated questions) and the set of references are the same as that given in figure \ref{fig:example}.}
\vspace{-5mm}
\label{fig:assignment_examples}
\end{figure*}

\noindent \textbf{1. Number Mismatch} - The average-metrics can fail to account for the mismatch between the number of predictions and the number of ground truth questions. 
As a concrete example, consider case 1 in Figure \ref{fig:error_case_examples}. The single prediction matches exactly with the first reference. Thus, the average $n$-gram match score of all the predictions (in this case, just one) will be perfect. However, there is only one prediction compared to three references. There are two references that the model failed to predict but the average-metrics cannot quantify this.
\setlength{\parskip}{1.0em}

\noindent \textbf{2. Reference Miscoverage} - The average-metrics fail to account for references that are not covered by the generated set. As a concrete example, consider case 2 in Figure \ref{fig:error_case_examples}. In this case there is no number mismatch. 
However, all the predictions are paraphrases of each other. As such, all of them match highly only with reference 1. Thus, the average $n$-gram match score of all predictions will be high. Nevertheless, the average-metrics remain oblivious to the failure of the model to generate anything close to reference 2 or reference 3.
\setlength{\parskip}{1.0em}

\noindent \citet{du2017identifying} proposed two evaluation strategies (``conservative" and ``liberal") to evaluate the full-system for paragraph-level multi-question generation but their schemes partly depend on aligning predictions and references based on their shared question-worthy sentence (if any). However, such a method is only possible for sentence-level or type-level granularity where each generated question is explicitly associated with a classified question-worthy sentence. In paragraph-level granularity the generated questions are not explicitly associated to any sentence. We propose a more model-agnostic and general evaluation strategy.  
\setlength{\parskip}{0em}

\subsection{Multi-metrics}
\label{sec:multi-metrics}
We propose a new evaluation scheme for comparing two sets of sequences to address the aforementioned limitations of existing strategies. First, we note that the one cause of reference miscoverage is the fact that multiple repetitive predictions can match with the same single reference. To solve this issue we decide to add a constraint. 
\setlength{\parskip}{1.0em}

\noindent \textbf{Constraint 1} - A prediction can be assigned to exactly one reference and exactly one prediction can be assigned to a reference. Only assigned pairs of prediction and reference can be scored. 
\setlength{\parskip}{1.0em}

\noindent That is, we first construct a mapping among predictions and references by assigning one to another. Next, we compute the matching scores (based on some sequence comparison metric) for \textit{only} the assigned pairs. Unlike before, we do not allow multiple predictions to match with the same reference or vice versa. Nevertheless, there can be multiple assignments that meet constraint 1. Thus, we set another constraint:
\setlength{\parskip}{1.0em}

\noindent \textbf{Constraint 2} -  Assume that there are $m$ predictions $(p_1, p_2, \dots, p_m)$, $n$ references $(r_1, r_2, \dots, r_n)$, and a sequence-level evaluation metric $M$ where $M(p_i, r_j)$ returns a score for how well prediction $p_i$ matches with reference $r_j$. We then choose a set $A$ of $k$ $(k = minimum(m,n))$ assignment pairs between predictions and references such that the ``overall match score" ($S$) of the assignments pairs are \textit{maximized}. Formally, the overall match score is computed as:  $S = \sum_{(p_i,r_j) \in A} M(p_i, r_j)$.
\setlength{\parskip}{1.0em}

\noindent The evaluation score $M$ can be any $n$-gram matching metric (e.g., BLEU or METEOR) or even newer metrics like BERTScore \cite{Zhang2020BERTScore}. We only consider metrics that return non-negative values. Combining constraint 1 and constraint 2, we get exactly the problem of optimal assignment. For example, we can treat predictions as workers, references as jobs and some $n$-gram match score of a particular pair of prediction and reference as the job-performance of the worker (prediction). For cases like these, the Hungarian algorithm \cite{kuhn1955hungarian}, which is a combinatorial optimization algorithm, was designed to find an optimal one-to-one assignment that maximizes the overall job-performance under the constraints. In our proposed evaluation scheme, we first use the Hungarian algorithm to find the $k$ assignments between predictions and references and also compute the overall match score $S$. In Figure \ref{fig:assignment_examples}, we show an example of the assignments made by the Hungarian algorithm based on METEOR (as $M$) on the prediction set and the reference set as given in Figure \ref{fig:example}. 
\setlength{\parskip}{0em}

Nevertheless, having this overall match score $S$ is not enough because the metric can ignore some references/predictions if there are more than $k$ references/predictions. As such, the problems with accounting number mismatch may remain. To resolve these issues, we use the overall score $S$ in an F-score-like framework. 

First, we compute a precision-like metric as $Pr = \frac{S}{m}$. $Pr$ reflects the overall degree of match that the predictions have with the references. The maixmum value of $S$ will be $c\cdot k$ where $c$ is the maximum value of evaluation $M$. Now, if there are too many predictions compared to ground truths then $m$ will be $\gg k$. High $m$ will reduce $Pr$ and mitigate the problem of over-prediction (one aspect of number mismatch).   

Second, we compute a recall-like metric as $Re = \frac{S}{n}$. $Re$ reflects the overall degree of match the references have with the predictions. That is, it reflects how well the references are covered by the predictions. If there are too few predictions compared to ground truths then $m$ will be $\ll n$; thus, $n$ will be $\gg k$. This will reduce $Re$ and mitigate the problem of under-prediction (another aspect of number mismatch). 

Finally, similar to F$_1$,  we compute an unweighted harmonic mean (Multi-$M$) of $Pr$ and $Re$: 
$$\textrm{Multi-}M = \frac{2\cdot Pr\cdot Re}{Pr+Re}$$

Here, in the name ``Multi-$M$", $M$ denotes the evaluation metric as mentioned before. $M$ can be any metric like BLEU or METEOR. Thus, we can have different corresponding variants of our set-level Multi-$M$ metrics as multi-BLEU or multi-METEOR depending on the metric $M$.

%% file: method.tex
\section{Generation Frameworks}
\label{sec:gen_frameworks}
As mentioned earlier, currently, state of the art in QG is achieved by pre-trained Seq2Seq models. Thus, we choose T5 \cite{raffel2020exploring} (a pre-trained Seq2Seq model) as the main model for question generation. For our specific task, there are several distinct strategies that we contrast utilizing T5. At a high level, some of these strategies involve generating one question at a time or generating a concatenated series of questions; generating at a document-level or generating at a sentence-level and collating the generations for every sentence. Overall, we consider two main factors of variation: (1) generation granularity and (2) generation mode. We describe them both below:

\subsection{Generation Granularity}
\label{sec:generation_granularity}
While we do solely focus on paragraph-level generation, that does not mean we have to \textit{directly} generate from the paragraph-level. Instead, we can also generate questions by focusing on a lower granularity (for example, sentences in the paragraph) and then collate the results in the end to have a paragraph-level output. Below, we discuss question generation in different levels of granularity. 
\setlength{\parskip}{1.0em}

\noindent \textbf{1. Paragraph-level} - 
In the paragraph-level granularity, we feed the whole paragraph to T5 and let it generate all the suitable questions from it directly. This model implicitly learns to find potential question-worthy areas (and just candidate answer-phrases) to generate questions about them.

\noindent \textbf{2. Sentence-level} - 
\label{sentence-level}
In the sentence-level granularity, we follow the strategy proposed by \citet{du2017identifying}. During training we train two models. One model is a question-worthiness classifier which classifies whether a sentence in a given paragraph is question-worthy or not. We describe the question-worthiness classifier in appendix \ref{sec:exp_details}. The other model is a seq2seq question generator which is trained to do sentence-level question generation. During inference, in the first step, we classify each sentence in the given paragraph as either question-worthy or not. In the second step, for every sentence classified as question-worthy we do sentence-level question generation using the trained question-generator. In the final step, we collate all the generated questions for every question-worthy sentences to represent the overall paragraph-level output. However, we integrate two notable differences from \citet{du2017identifying}: 1. We use T5 instead of an RNN-based question-generator; 2. Instead of using just a sentence as input for the sentence-level question generator, we use the whole paragraph (after flattening it into a single sequence of tokens) as input. We however ``highlight" the sentence in the paragraph to make the generator focus on that highlighted sentence for sentence-level generation. Highlighting is done by adding a special token \verb+<hl>+ at the beginning of the sentence, and a special token \verb+</hl>+ at its end. This was done because sometimes the surrounding context of a sentence is necessary to generate the ground truth questions \cite{zhao2018paragraph}. 
\begin{figure}[t]
    \centering
    \includegraphics[scale=0.15]{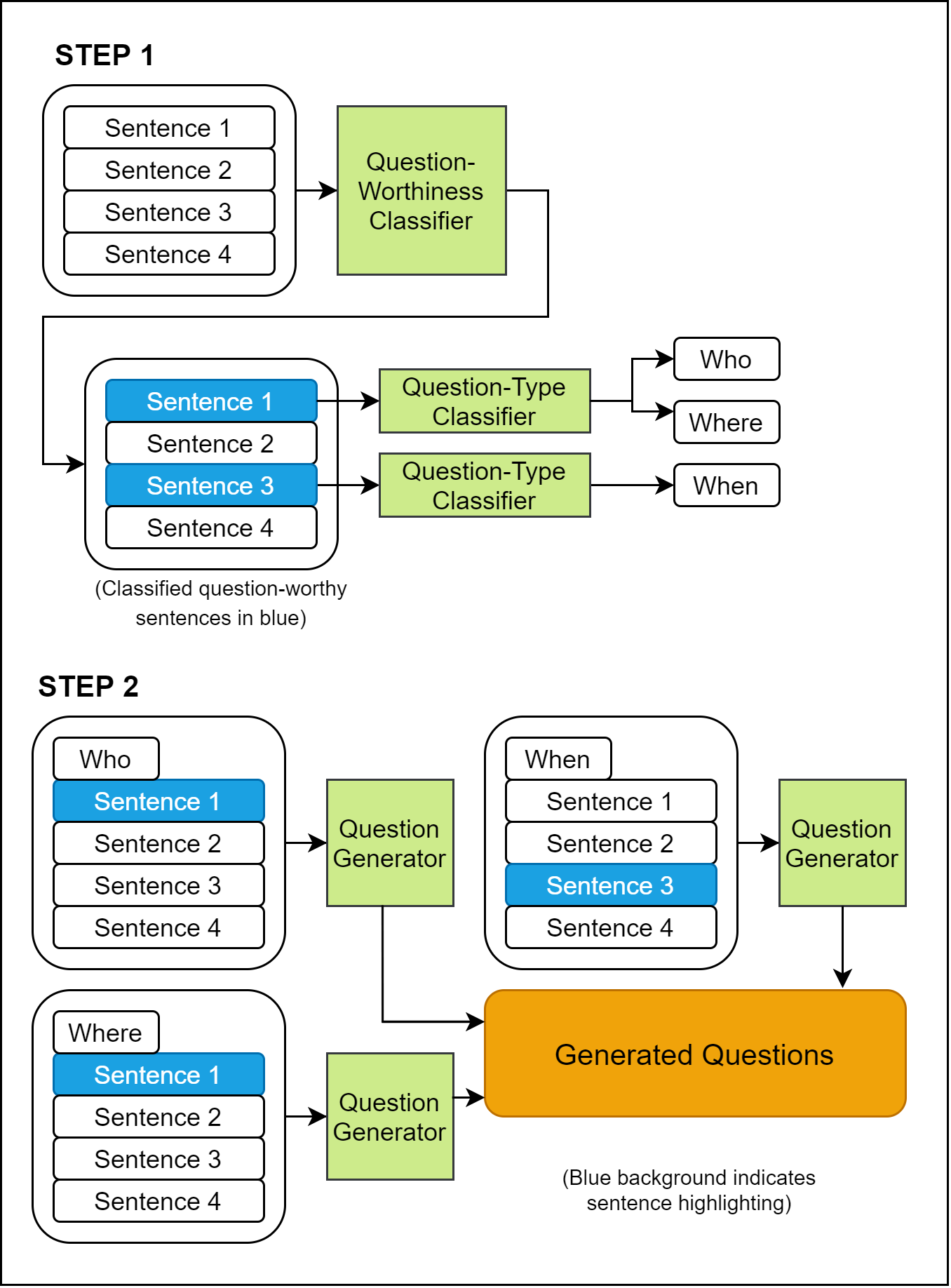}
    \caption{Framework for QG at a type-level granularity during inference. Step 1 shows question-worthiness classification followed by question-type prediction from the classified question-worthy sentences (highlighted in blue). Step 2 shows generating type-conditioned questions for each predicted question-type and its corresponding question-worthy sentence (highlighted in blue). Generation at a sentence-level granularity works similarly but excluding the question-type classifier (step 1) and type-conditioning (step 2).}
    \label{fig:type-level generation}
\end{figure}
\noindent \textbf{3. Type-level} - 
In type-level generation, during training, we train three separate models. As in {\em sentence-level}, we train a question-worthiness classifier and a T5-based question-generator. However, in addition, we also train a question-type classifier which predicts all \textit{types} (e.g., who, where, how etc.) of questions that are worthy to be asked from the question-worthy sentences. We frame the question-type prediction task as a multi-label sentence classification problem. We consider question type labels $\in \{\textrm{who}, \textrm{when}, \textrm{where}, \textrm{what}, \textrm{why}, \textrm{which}, \textrm{how}, \\ \textrm{quantity}, \textrm{other}\}$. We describe the question-type classifier in appendix \ref{sec:exp_details}. Different from {\em sentence-level}, we train the T5 sentence-level question-generator to be conditioned by the question-type. Essentially, we prepend a special token indicating the ground-truth question type to the input of the sentence-level question-generator during training. As such, the generator learns to condition its generation on the question-type as specified by the prepended token. A similar sentence-highlighting strategy is used as before for the generator input.
\setlength{\parskip}{0.0em}

During inference, first, we classify each sentence in a given paragraph to check whether it is \textit{question-worthy} or not. Second, for every classified question-worthy sentence we predict all the types of questions that are worthy to be asked from that sentence. Third, for every question-worthy sentence and for every question type appropriate to be asked from that sentence, we perform a type-conditioned sentence-level question generation for that question-type (conditioned by preprending) and that sentence. Finally, all the generated questions for each sentence and its question types, can be collated together to have an output set of questions for the overall paragraph. Figure \ref{fig:type-level generation} shows the pipeline for type-level generation during inference. 


Similar to us, \citet{wu2020question} used type-driven generation in an answer-agnostic setup but they focus on sentence-level single question generation. Moreover, they frame question type prediction as a multi-class classification problem (question types for a sentence are considered to be mutually exclusive). While \citet{wu2020question} proposed to select top $K$ most probable {\em question types} to generate multiple questions for multiple types for the same sentence, the model is still dependent on the hyperparameter $K$. Instead, in our approach, we simply let the classification model decide which and how many types to predict by framing question-type prediction as a \textit{multi-label} classification problem. 

\subsection{Generation Mode}
\label{sec:generation_mode}
Besides granularity, the generation multiplicity (or mode) is another factor to consider. We consider two different modes of generation, which we discuss  below.
\setlength{\parskip}{1.0em}

\noindent \textbf{1. One2One Generation} - 
In one2one generation, we train the question-generator model to maximize the likelihood for a single ground truth question for the sample input (of whatever granularity). To have generation of multiple questions in an one2one setting, we can use beam search or multiple sampling-based decoding. We can also generate multiple questions by generating at a lower granularity and then collating the results. 

\noindent \textbf{2. One2Many Generation} - 
In one2many generation settings, we train the question-generation model to maximize the likelihood of the \textit{concatenation} of all the questions that are asked (in the ground truth set) for the given sample input of a specific granularity. An one2many model can generate multiple questions even when using greedy decoding. A key benefit of the one2many mode is that we can allow the model itself to determine how many questions to generate. In contrast, in the one2one setting, we have to manually decide how many questions we want to generate (using beam search or sampling techniques). Another benefit of one2many mode is that the generation of one question can be informed by prior generated questions for the given input. 
\setlength{\parskip}{0.0em}

A similar one2many mode of generation was proposed for keyphrase generation \cite{yuan-etal-2020-one}. For question generation, in particular, \citet{lopez2021transformerqg} used the one2many mode of generation with pre-trained Transformer models. 
Different from any prior works, we explore combinations of one2many generation in sentence and type-level granularity considering that there could be multiple questions that can be asked even for a single sentence or a question type.

For training one2many models, we simply prepare the training ground truths by concatenating all the questions (for the given granularity of input) while using a special token \verb+<sep>+ as a delimiter. The concatenation of questions is done in the order (first to last) of the occurrence of their corresponding question-worthy sentences.  
\setlength{\parskip}{0.5em}

\noindent Overall, there are six ways of combining the above factors (granularity and mode), and thus a total of six possible strategies. We compare all six.  
\setlength{\parskip}{0em}

%% file: rank_and_select.tex
\section{Ranking and Selection}
\label{sec:rank_and_select}
When using one2one generation mode, we have to also decide on a method to generate multiple questions, particularly when generating directly at the paragraph-level granularity. 
Thus, for one2one mode, we employ an overgenerate-and-rank strategy. First we overgenerate multiple questions (the overgeneration method is discussed in section \ref{sec:overgeneration}). Second, we rank the generations using some ranking method. Third, we select some $k$ highest ranked generations. For ranking and selection, we consider the following methods:
\setlength{\parskip}{1.0em}

\noindent \textbf{1. Rand@5}: In this method, we randomly select $5$ generations. The average number of questions per paragraph is also $5$ in the dataset that we use.

\noindent \textbf{2. Top@1}: In this method we only generate and select a single greedy-decoded question. Thus, in this case, there will be only one generated question per input at a specific granularity-level. For example, in case of paragraph-level granularity there will be only one generation overall and in case of sentence-level granularity there can be as many generations as there are question-worthy sentences.

\noindent \textbf{3. Rank@5}: In this method, we rank the questions based on their answerability probability and then select the top $5$ questions. We use a neural question-answering model to predict the answerability probability for each question. In particular, we use an ELECTRA-large model \cite{Clark2020ELECTRA} that was trained on SQuAD2.0 \cite{rajpurkar2018know} to not only extract an answer span but also to decide whether the question is answerable from the given context.\footnote{\url{https://huggingface.co/ahotrod/electra_large_discriminator_squad2_512}} We also filter any question with less than $0.5$ answerability probability.  
\setlength{\parskip}{0em}

\subsection{Overgeneration}
\label{sec:overgeneration}
For overgenerating multiple ($K$) questions in one2one methods, we generate one question using greedy decoded for the given granularity level. Rest of the $K$ generations at the given granularity level is decoded using multiple runs of Nucleus Sampling \cite{Holtzman2020The} with a top-p value of $0.9$. We set $K=20$ for paragraph-level granularity, $K=10$ for sentence-level granularity, and $K=5$ for type-level granularity. The $K$-values were chosen arbitrarily for most part. We decrease the $K$-value for lower granularities because the overall generation can be multiple times higher than $K$ depending on how many question-worthy sentences or how many question-types there are.

%% file: experiments.tex
\section{Experiments}
\label{sec:experiments}
\begin{table}[t]
\small
\centering
\def\arraystretch{1.2}
\begin{tabular}{  l | r | r | r | r}
\toprule
\textbf{Model} & \textbf{Prec.} & \textbf{Rec.} & \textbf{F$_1$} & \textbf{Acc.}\\
\hline
\citet{du2017identifying} & 73.15 & \textbf{89.29} & 80.42 & 72.52\\
ELECTRA CL & \textbf{76.19} & 74.02 & 75.09 & 70.11\\
ELECTRA HC & 75.88 & 88.24 & \textbf{81.59} & \textbf{75.77}\\
\bottomrule
\end{tabular}
\caption{Question-worthiness classification results} 
\label{table:qw_classification}
\end{table}

\begin{table}[t]
\small
\centering
\def\arraystretch{1.2}
\begin{tabular}{  l | r | r | r }
\toprule
\textbf{Type} & \textbf{Prec.} & \textbf{Rec.} & \textbf{F$_1$}\\
\hline
Who & 39.64 & 88.75 & 54.80\\
When & 31.85 & 78.63 & 45.34\\
Where & 21.62 & 79.05 & 33.96\\
What & 83.59 & 49.11 & 61.87\\
Quantity & 56.72 & 88.65 & 69.18\\
How & 10.97 & 65.53 & 18.79\\
Why & 09.12 & 64.84 & 15.99\\
Which & 20.86 & 64.78 & 31.55\\
Others & 02.35 & 32.26 & 04.37\\
\bottomrule
\end{tabular}
\caption{Question-type classification results} 
\label{table:qt_classification}
\end{table}

In this section, we discuss the details of our experimental models, datasets, evaluation, and results. We use an ELECTRA-large-based classifier \cite{Clark2020ELECTRA} for question-worthiness classification and question-type classification. More details on the classifier architectures are in section \ref{sec:exp_details}. Hyperparameter details are in section \ref{sec:classifier_hyperparameters} and \ref{sec:generator_hyperparameters}.
\subsection{Classifier Architecture}
\label{sec:exp_details}
\noindent \textbf{Question-type Classification:} For Question-Type classification, we use ELECTRA-large as a multi-label sentence classifier. We transform the final representation of the \verb+CLS+ token using two layers with a GELU activation \cite{hendrycks2016gelu} in between for classification.
\setlength{\parskip}{1em}

\noindent \textbf{Question-worthiness Classification}: For Question-worthiness classification we try two distinct approaches: (1) ELECTRA CL and (2) ELECTRA HC. We discover that ELECTRA HC outperforms ELECTRA CL and so only ELECTRA HC is used in the main experiments (Table \ref{table:SQuAD main results},\ref{table:SQuAD secondary results}). Table \ref{table:qw_classification} also shows the result of ELECTRA HC. 
We describe both methods below:

\noindent\textbf{1. ELECTRA CL:} In this approach we simply use ELECTRA-large as a sentence-level binary classifier similar to how we use it as a multi-label classifier for question-type classification.

\noindent\textbf{2. ELECTRA HC:} ELECTRA HC takes a hierarchical approach towards classification. It uses a similar framework as used by \citet{du2017identifying}. First, we encode each sentence in a paragraph into a single vector (sentence-vector) using a sentence encoder (as a result, we will have a sequence of sentence-vectors representing the paragraph). Second, the sequence of vectors are contextualized by a BiLSTM \cite{hochreiter1997lstm,graves2005bilstm}. Third, we apply a binary classifier (two layers with a GELU activation in between) for each of the sentence-vectors in the sequence. In this strategy the classification of question-worthiness of each sentence is informed by the context of the surrounding sentences in the paragraph. Different from \citet{du2017identifying}, we use ELECTRA-large, a more modern model, for sentence encoding. We treat the final representation of the \verb+CLS+ token of ELECTRA as the encoded sentence vector for each sentence.
\setlength{\parskip}{0em}

\subsection{Question-type Determination}
\label{sec:type_determination}
For most of the part, we determine the question-type of a question based on whether the question-type word is present in the question. For example, if ``who" is present in a question, its question-type is determined to be ``who". Sometimes the question-type word occur in the middle of the question so we did not restricted ourselves to only checking the first question word (several examples of such cases can be observed in table \ref{table:samples1} reference questions). This rule is not foolproof, but generally works in the dataset that we use. There are, however, multiple exceptions to the general rule mentioned above. If ``whose" or ``whom" are present in the question, the question-type is still determined as ``who". If ``how much" or ``how many" is present in the question then the question-type is determined to be ``quantity" instead of ``how". We do this because general ``how" questions are of a different breed (asking for a process) than questions asking ``how much"/``how many" (generally asking for some quantity or intensity). Presence of the words ``quantity" or ``other" in a question do not determine the question-type to be ``quantity" or ``other". Any questions whose type remains undetermined by the above rules are determined as of type ``other". Questions with answers ``yes" or ``no" are also determined to be of type ``other". We do not keep a separate type for boolean questions (yes and no questions) because there are very few instances of this type in the dataset that we use. 

\subsection{Classifier Hyperparameters}
\label{sec:classifier_hyperparameters}
All ELECTRA-based classifiers use two layers on top of the final representation of the \verb+CLS+ token. There is a GELU activation function \cite{hendrycks2016gelu} in between. The first layer has as many neurons as in the ELECTRA hidden state (except when we use ELECTRA HC. For ELECTRA HC the number of neurons of this layer is the same as that of BiLSTM hidden size). The last layer has a sigmoid activation function. In case of question-worthiness classification (binary classification) the last layer have a single neuron. In case of question-type classification, the last layer have as many neurons as there are question-type labels. The total hidden size of the BiLSTM hidden state (forward and backward combined), as used in ELECTRA HC, is $300$. Some of the general hyperparameters used for all classifiers are a weight-decay of $0.01$, a maximum gradient normalization clipping of $5$, a maximum sequence length of $512$, a batch size of $64$, a maximum number of epochs of $30$, and an early stop patience of $2$ (the training is terminated when the loss does not reduce for two consecutive epochs). We also use RAdam \cite{Liu2020On} as the optimizer. For all classifiers, the learning rate is tuned using grid-search among the options $\{0.001, 0.0002, 0.0001, 0.00002\}$. $0.00002$ was chosen for each. Label weights were used for question-type classification. The label weight for a particular label were determined as the total number of negative instances divided by the total number of positive instances of that label. Model selection and hyperparameter selection are done based on the best validation loss.

\subsection{Generator Hyperparameters}
\label{sec:generator_hyperparameters}
Some of the shared hyperparameters of the T5-based question generator models are a batch size of $128$, a weight decay of $0$, a maximum number of epochs as $20$, a maximum sequence length of $512$, a maximum gradient normalization clipping of $5$, and an early stop patience of $2$ (the training is terminated when the loss does not reduce for two consecutive epochs). We also use SM3 \cite{rohan2019memory} as the optimizer. Greedy decoding is used for generation in one2many mode. One2one modes utilize overgenerate-and-rank methods which are discussed earlier. The learning rate is tuned using grid-search among the options $\{1.0, 0.1, 0.01, 0.001\}$ for all generators separately. We limit the maximum epochs to $10$ for each hyperparameter tuning trial. A learning rate of $0.1$ was chosen for paragraph-level one2many generation and a learning rate of $0.01$ was chosen for the rest. Learning rate (lr) warmup  as follows (based on the recommended procedure for SM3 \footnote{\url{https://github.com/google-research/google-research/tree/master/sm3}}):
\begin{equation}
    lr_s = lr_0 \cdot min(1,(s/w)^2)
\end{equation}
$lr_s$ indicates the learning rate at step $s$. The inital learning rate ($lr_0$) is whatever is chosen after hyperparameter tuning. $s$ indicates the current update step number. $w$ indicates total warmup steps (set as $2000$). Model selection and hyperparameter selection are done based on the best validation loss.

\subsection{Dataset}
We perform all our experiments on the processed SQuAD1.1 \cite{rajpurkar2016squad} split as provided by \citet{du2017learning}\footnote{\url{https://github.com/xinyadu/nqg/tree/master/data/processed}} for question generation. For question-worthiness classification, all the sentences from all the paragraphs in the dataset are input samples. A sample input sentence is considered question-worthy if that sentence has a corresponding ground-truth question in the dataset. For question-type classification, only question-worthy sentences are sample inputs. For each sample, its question-type labels are the question-types of all the questions associated to that sample sentence. The question-type of a question is determined based on some heuristic rules that are detailed in appendix \ref{sec:type_determination}. We maintain the original train-development-test split (as provided by \citet{du2017learning}) for all experiments. 

\begin{table*}[t]
\small
\centering
\def\arraystretch{1.2}
\begin{tabular}{  l l | R{1cm} R{1cm} R{1cm} R{1cm} r r r r} 
\toprule
\textbf{Granularity} & \textbf{Generation Mode} & \textbf{Multi-BLEU4} & \textbf{Multi-MTR} &\textbf{Multi-R-L} &\textbf{Multi-qBLEU1} & \textbf{BLEU4} & \textbf{MTR} & \textbf{R-L} & \textbf{qBLEU1}\\
\hline
\multicolumn{2}{l|}{\citet{du2017identifying}} & --- & --- & --- & --- & 12.28 & 16.62 & 39.75 & ---\\
\multicolumn{2}{l|}{\citet{lopez2021transformerqg}} & --- & --- & --- & --- & 8.26 & 21.2 & \textbf{44.38} & --\\
\hline
Paragraph & one2one (Rand@5) & 5.14	& 15.78	& 28.79	& 28.78	 & 7.78	& 20.52	& 37.67 & 43.50\\
Paragraph & one2one (Top@1) & 3.73 & 9.10 & 16.94 & 16.58 & \textbf{12.68} & 23.58 & 44.12 & \textbf{49.27}\\
Paragraph & one2one (Rank@5) & 6.78	& 17.07	& 30.17	& 30.38	& 11.21	& 23.71	& 41.69 & 47.98\\
\cdashline{1-10}
Paragraph & one2many & 6.25 & 15.55 & 28.97 & 28.48 & 11.15 & 22.12 & 42.41 & 47.58\\
\hline
Sentence & one2one (Rand@5) & 5.37 & 15.85 & 28.97 & 29.24	& 7.85 & 20.48 & 37.86 & 43.88\\
Sentence & one2one (Top@1) & 6.97 & 16.56 & 30.02 & 29.96 & 11.91 & 23.20 & 42.81 & 48.61\\
Sentence & one2one (Rank@5) & 6.77 & 16.70 & 29.50 & 29.76 & 11.22 & \textbf{23.72} & 41.51 & 47.86\\ 
\cdashline{1-10}
Sentence & one2many & \textbf{7.80} & \textbf{17.93} & \textbf{32.08} & \textbf{32.08}	& 11.64 & 22.65 & 41.96 & 47.60 \\
\hline
Type & one2one (Rand@5) & 4.86	& 15.34	& 28.07	& 28.15	& 7.15 & 19.82 & 36.37	 & 42.36\\
Type & one2one (Top@1) & 7.01 & 16.03 & 29.06 & 28.90 & 12.30	& 23.27	& 42.78 & 48.55\\
Type & one2one (Rank@5) & 6.43 & 16.42	& 29.00 & 29.13 & 10.45 & 23.08 & 40.35 & 46.57\\
\cdashline{1-10}
Type & one2many & 7.57	& 16.79	& 30.05	& 30.11	& 12.67	& 23.4	& 42.65 & 48.54\\
\bottomrule
\end{tabular}
\caption{Multi-metrics and average metrics performance on paragraph-level multi-question generation on SQuAD2.0 (split by \cite{du2017learning}) using different granularity-levels and generation modes.} 
\label{table:SQuAD main results}
\end{table*}

\begin{table}[t]
\small
\centering
\def\arraystretch{1.2}
\begin{tabular}{  l l | R{1cm} r} 
\toprule
\textbf{Granularity} & \textbf{Generation Mode} & \textbf{Self-BLEU2} & \textbf{car. diff}\\
\hline
Paragraph & one2one (Rand@5) & 66.18 & -0.11\\
Paragraph & one2one (Top@1) & 0	& 3.89\\
Paragraph & one2one (Rank@5) & 49.25 & -0.11\\
\cdashline{1-4}
Paragraph & one2many & 38.82 & 1.45\\
\hline
Sentence & one2one (Rand@5) & 72.39 & -0.10 \\
Sentence & one2one (Top@1)& 17.84 & 1.78\\
Sentence & one2one (Rank@5)& 52.72 & -0.10\\
\cdashline{1-4}
Sentence & one2many & 34.91 & 0.39\\
\hline
Type & one2one (Rand@5)& 71.32 & -0.10\\
Type & one2one (Top@1)& 16.43 & 1.99\\
Type & one2one (Rank@5)& 50.63	& -0.09\\
\cdashline{1-4}
Type & one2many & 17.04 & 1.76\\
\bottomrule
\end{tabular}
\caption{Self-BLEU and cardinality difference (car. diff) for QG on \citet{du2017learning} split using different granularity-levels and generation modes.}
\vspace{-1em}
\label{table:SQuAD secondary results}
\end{table}

\begin{table*}[t]
\small
\centering
\def\arraystretch{1.2}
\begin{tabular}{l}
\toprule
\textbf{Example \# 1}\\
\textbf{Generated Questions:}\\
1. what were the early courses in in the college of science?\\
2. when was the college of engineering established?\\
\textbf{Reference Questions:}\\
1. how many bs level degrees are offered in the college of engineering at notre dame?\\
2. in what year was the college of engineering at notre dame formed?\\
3. before the creation of the college of engineering similar studies were carried out at which notre dame college?\\
4. how many departments are within the stinson-remick hall of engineering?\\
5. the college of science began to offer civil engineering courses beginning at what time at notre dame?\\
\textbf{BLEU4:} $40.34$; \textbf{Multi-BLEU4:} $13.26$; \textbf{METEOR:} $22.06$; \textbf{Multi-METEOR:} $11.81$;\\
\textbf{ROUGE-L:} $42.38$; \textbf{Multi-ROUGE-L:} $22.91$; \textbf{Q-BLEU1:} $40.26$; \textbf{Multi-Q-BLEU1:} $17.01$\\
\midrule
\textbf{Example \# 2}\\
\textbf{Generated Questions:}\\
1. when was theodore m. hesburgh library completed?\\
2. what is the library system of the university divided between?\\
3. what is the name of the library that houses the main collection of books?\\
4. what is the word of life mural known as?\\
5. what does the word of life mural appear to make?\\
6. who designed the word of life mural?\\
\textbf{Reference Questions:}\\
1. how many stories tall is the main library at notre dame?\\
2. what is the name of the main library at notre dame?\\
3. in what year was the theodore m. hesburgh library at notre dame finished?\\
4. which artist created the mural on the theodore m. hesburgh library?\\
5. what is a common name to reference the mural created by millard sheets at notre dame?\\
\textbf{BLEU4:} $10.65$; \textbf{Multi-BLEU4:} $11.38$; \textbf{METEOR:} $17.25$; \textbf{Multi-METEOR:} $15.04$;\\
\textbf{ROUGE-L:} $40.15$; \textbf{Multi-ROUGE-L:} $33.6$; \textbf{Q-BLEU1:} $36.44$; \textbf{Multi-Q-BLEU1:} $27.72$\\
\midrule
\textbf{Example \# 3}\\
\textbf{Generated Questions:}\\
1. what do most people with dogs describe their pet as?\\
2. what does a study of conversations in dog-human families show?\\
3. what is the popular reconceptualization of the dog-human family as a pack?\\
4. what is a dominance model of dog-human relationships promoted by some dog trainers?\\

\textbf{Reference Questions:}\\
1. how do most people describe the relationship with their dogs?\\
2. what television show uses a dominance model of dog and human relationships?\\
3. most people today describe their dogs as what?\\
4. what tv show promotes a dominance model for the relationships people have with their dogs?\\
\textbf{BLEU4:} $05.56$; \textbf{Multi-BLEU4:} $05.56$; \textbf{METEOR:} $24.28$; \textbf{Multi-METEOR:} $21.21$;\\
\textbf{ROUGE-L:} $37.13$; \textbf{Multi-ROUGE-L:} $32.43$; \textbf{Q-BLEU1:} $41.97$; \textbf{Multi-Q-BLEU1:} $37.10$\\
\bottomrule
\end{tabular}
\caption{SQuAD1.1 examples with generations from T5 sentence-level one2many. Comparison between different evaluation metrics are presented.} 
\label{table:samples1}
\end{table*}

\begin{table*}[t]
\small
\centering
\def\arraystretch{1.2}
\begin{tabular}{l}
\toprule
\textbf{Example \# 1}\\
\textbf{Generated Questions:}\\
1. what is the name of the oldest building on campus?\\
\textbf{Reference Questions:}\\
1. where is the headquarters of the congregation of the holy cross?\\
2. what is the primary seminary of the congregation of the holy cross?\\
3. what is the oldest structure at notre dame?\\
4. what individuals live at fatima house at notre dame?\\
5. which prize did frederick buechner create?\\
\textbf{BLEU4:} $0$; \textbf{Multi-BLEU4:} $0$; \textbf{METEOR:} $17.58$; \textbf{Multi-METEOR:} $05.86$;\\
\textbf{ROUGE-L:} $50$; \textbf{Multi-ROUGE-L:} $15.12$; \textbf{Q-BLEU1:} $43.00$; \textbf{Multi-Q-BLEU1:} $12.07$\\
\midrule
\textbf{Example \# 2}\\
\textbf{Generated Questions:}\\
1. what is the name of the college of engineering?\\
\textbf{Reference Questions:}\\
1. how many bs level degrees are offered in the college of engineering at notre dame?\\
2. in what year was the college of engineering at notre dame formed?\\
3. before the creation of the college of engineering similar studies were carried out at which notre dame college?\\
4. how many departments are within the stinson-remick hall of engineering?\\
5. the college of science began to offer civil engineering courses beginning at what time at notre dame?\\
\textbf{BLEU4:} $43.44$; \textbf{Multi-BLEU4:} $07.54$; \textbf{METEOR:} $24.33$; \textbf{Multi-METEOR:} $08.11$;\\
\textbf{ROUGE-L:} $49.23$; \textbf{Multi-ROUGE-L:} $15.47$; \textbf{Q-BLEU1:} $47.69$; \textbf{Multi-Q-BLEU1:} $12.52$\\
\bottomrule
\end{tabular}
\caption{SQuAD1.1 examples with generations from T5 paragraph-level one2one (Top@1). Comparison between different evaluation metrics are presented.} 
\label{table:samples2}
\end{table*}

\subsection{Evaluation}
We use standard precision, recall, F1 measures for the classification tasks. For question generation, we use different instances of multi-metrics (discussed in $\S$\ref{sec:multi-metrics}): Multi-BLEU4, Multi-METEOR (Multi-MTR), Multi-ROUGE-L (Multi-R-L), and Multi-qBLEU1. We also report the average BLEU4, METEOR (MTR), ROUGE-L (R-L), and qBLEU1 metrics. In Table \ref{table:SQuAD secondary results}, we use self-BLEU2 \cite{zhu2018texygen} on the prediction set to show prediction diversity (Lower self-BLEU means higher diversity). We also show the difference of the number of predictions from the number of ground truth questions. We refer to this metric as cardinality difference (car. diff). This shows the degree of number mismatch. We use nlg-eval \cite{sharma2017nlgeval} for $n$-gram match scores. For Q-BLEU1 we use the same parameters as recommended for Q-BLEU1 on SQuAD by the original paper (See table 5 in \cite{nema2018towards}).

\subsection{Results} 
Table \ref{table:qw_classification} compares the performance of ELECTRA HC and ELECTRA CL. As can be seen, ELECTRA HC outperforms ELECTRA CL by a significant margin and obtains the overall best performance on recall, F1, and accuracy. In Table \ref{table:qt_classification}, we show the performance for each question-type labels. While the type classification performance is low, it is on par with results obtained by prior methods \cite{wu2020question} (although the results are not strictly comparable due to differences in question-types).

In Table \ref{table:SQuAD main results}, we show the main results of question generation. Among the one2one ranking methods, Top@1 seems to work best at sentence-level or type-level granularities but it causes higher magnitude cardinality difference (as shown in Table \ref{table:SQuAD secondary results}). The multi-metrics performance for Top@1 paragraph-level is poor because it can only predict one question for the whole paragraph causing high number mismatch. Rank@5 works better than Rand@5 for any granularity level, which makes sense given that Rank@5 takes answerability into account. 
Among generation modes, one2many generally works better than one2one ranking when using sentence-level or type-level granularity. Among granularity-levels, sentence-level seems to generally perform the best on multi-metrics. Type-level generation does not offer much further benefit. Type-level generation also causes higher magnitude cardinality difference (Table \ref{table:SQuAD secondary results}) although they get lower self-BLEU (higher diversity) (Table \ref{table:SQuAD secondary results}). However, in some cases, lower self-BLEU can result from under-prediction (having fewer predictions reduces the chances of $n$-gram overlaps among each other); thus, lower self-BLEU is not always good.

\section{On the Efficacy of Multi-metrics}
\label{sec:efficacy}
We motivate the efficacy of multi-metrics on two different grounds. 

First, we emphasize the theoretically established virtues of multi-metrics in terms of its ability to account for failure in reference miscoverage and also, number mismatch (in $\S$\ref{sec:multi_evaluation}). 

Second, we show a concrete instance where average metrics are high despite a critical failure of the model, whereas multi-metrics are low as it should be. This can be observed in Table \ref{table:SQuAD main results} in the case of paragraph-level one2one generation with Top@1 ranking. Here, the model can only predict one question at most, whereas there are five references on average. Thus, we see a high cardinality difference in Table \ref{table:SQuAD secondary results} for this case. Despite this, the average $n$-gram metrics obtain very high scores for this model. These metrics are completely insensitive to this model failure. Multi-metrics, on the other hand, can take this failure into account. Thus, we observe multi-metrics assign it the lowest scores. 


We also present some concrete examples (of generated set of questions and ground truth set of questions or references) along with their corresponding metrics (both average-based n-gram match metrics and multi-metrics) in table \ref{table:samples1} and \ref{table:samples2}. In table \ref{table:samples1} example \# 1, we find a degree of number mismatch. There are only two predictions whereas there are five references. As expected, we find a quite a bit of difference between multi-metrics and average-metrics here because multi-metrics is penalized heavily because of number mismatch and average-metric is not penalized as much. In example \# 2 and \# 3 from table \ref{table:samples1}, we find that the multi-metrics are much closer to the average metrics because the number of predictions are close to the number of references; and furthermore, there are no issues with paraphrases in predictions. On the other hand, in the examples in table \ref{table:samples2}, we again observe amplified difference between multi-metrics and average-metrics given the high degree of number mismatch that the average-metrics fail to take into account.

%% file: related_work.tex
\section{Related Work}
\label{sec:related}
\textbf{Assignment-based Evaluations -} \citet{rus2012comparison} proposed an optimal-matching-based method for embedding-based text similarity measure but not in the context of comparing \textit{sets} of sequences in an F$_1$-like framework. Similarly, several text evaluation approaches \cite{kusner2015word, chow2019wmdo, clark2019sentence, zhao2019moverscore} used earth mover's distance \cite{rubner1998metric} whereas \citet{Zhang2020BERTScore} used greedy-matching. 
Similar to us, \citet{sejr2020evaluating} proposed an evaluation for QG but using greedy matching (instead of optimal assignment) which allows multiple predictions to match with a single reference and vice versa.

\noindent\textbf{Question Generation-} \citet{du2017learning} presented one of the earliest works on answer-agnostic neural QG. 
There were also several early answer-aware QG approaches \cite{yuan2017machine,zhou2017neural}. Several works took joint-training or multi-task approaches to train both question answering and QG \cite{duan2017question, song2017unified,duyu2017question,wang2017joint}. \citet{du2017identifying} proposed QG in the sentence-level granularity. \citet{subramanian2018neural} generated questions based on detected keyphrases. Similarly, \citet{wang2019multi-agent} used a multi-agent communication framework to first identify question-worthy phrases and then generate questions with their assistance. \citet{zhao2018paragraph} used maxout pointer and gated self-attention to exploit paragraph-level information for QG. \citet{scialom2019self} used Transformer-based approaches for answer-agnostic QG. Multiple works \cite{fan2018question-type, sun2018answer, hu2018aspect,zhou2019questiontype,wu2020question} utilized question-words information or a question-type driven framework for different variants of QG. Newer approaches \cite{chan2019bert,chan2019recurrent,dong2019unified,varanasi2020copybert,qi2020prophetnet,lopez2021transformerqg} used pre-trained Transformer models.

%% file: conclusion.tex
\section{Conclusion}
\label{sec:conclusion}
We proposed a new evaluation method (multi-metrics) to evaluate multi-question generation. We motivate the evaluation theoretically and also empirically in terms of the contrast discussed in $\S$\ref{sec:efficacy}. Using both new and old evaluations, we also empirically compare combinations of various strategies for paragraph-level multi-question generation under a common framework. Our results show that using factorized sentence-level generation in one2many mode is better than directly generating from paragraph-level even when using powerful pre-trained models.

%% file: main.bbl
\begin{thebibliography}{57}
\expandafter\ifx\csname natexlab\endcsname\relax\def\natexlab#1{#1}\fi

\bibitem[{Anil et~al.(2019)Anil, Gupta, Koren, and Singer}]{rohan2019memory}
Rohan Anil, Vineet Gupta, Tomer Koren, and Yoram Singer. 2019.
\newblock \href
  {https://proceedings.neurips.cc/paper/2019/file/8f1fa0193ca2b5d2fa0695827d8270e9-Paper.pdf}
  {Memory efficient adaptive optimization}.
\newblock In \emph{Advances in Neural Information Processing Systems},
  volume~32. Curran Associates, Inc.

\bibitem[{Banerjee and Lavie(2005)}]{banerjee-lavie-2005-meteor}
Satanjeev Banerjee and Alon Lavie. 2005.
\newblock \href {https://aclanthology.org/W05-0909} {{METEOR}: An automatic
  metric for {MT} evaluation with improved correlation with human judgments}.
\newblock In \emph{Proceedings of the {ACL} Workshop on Intrinsic and Extrinsic
  Evaluation Measures for Machine Translation and/or Summarization}, pages
  65--72, Ann Arbor, Michigan. Association for Computational Linguistics.

\bibitem[{Chan and Fan(2019{\natexlab{a}})}]{chan2019bert}
Ying-Hong Chan and Yao-Chung Fan. 2019{\natexlab{a}}.
\newblock \href {https://doi.org/10.18653/v1/W19-8624} {{BERT} for question
  generation}.
\newblock In \emph{Proceedings of the 12th International Conference on Natural
  Language Generation}, pages 173--177, Tokyo, Japan. Association for
  Computational Linguistics.

\bibitem[{Chan and Fan(2019{\natexlab{b}})}]{chan2019recurrent}
Ying-Hong Chan and Yao-Chung Fan. 2019{\natexlab{b}}.
\newblock \href {https://doi.org/10.18653/v1/D19-5821} {A recurrent
  {BERT}-based model for question generation}.
\newblock In \emph{Proceedings of the 2nd Workshop on Machine Reading for
  Question Answering}, pages 154--162, Hong Kong, China. Association for
  Computational Linguistics.

\bibitem[{Chow et~al.(2019)Chow, Specia, and Madhyastha}]{chow2019wmdo}
Julian Chow, Lucia Specia, and Pranava Madhyastha. 2019.
\newblock \href {https://doi.org/10.18653/v1/W19-5356} {{WMDO}: Fluency-based
  word mover{'}s distance for machine translation evaluation}.
\newblock In \emph{Proceedings of the Fourth Conference on Machine Translation
  (Volume 2: Shared Task Papers, Day 1)}, pages 494--500, Florence, Italy.
  Association for Computational Linguistics.

\bibitem[{Clark et~al.(2019)Clark, Celikyilmaz, and Smith}]{clark2019sentence}
Elizabeth Clark, Asli Celikyilmaz, and Noah~A. Smith. 2019.
\newblock \href {https://doi.org/10.18653/v1/P19-1264} {Sentence mover{'}s
  similarity: Automatic evaluation for multi-sentence texts}.
\newblock In \emph{Proceedings of the 57th Annual Meeting of the Association
  for Computational Linguistics}, pages 2748--2760, Florence, Italy.
  Association for Computational Linguistics.

\bibitem[{Clark et~al.(2020)Clark, Luong, Le, and Manning}]{Clark2020ELECTRA}
Kevin Clark, Minh-Thang Luong, Quoc~V. Le, and Christopher~D. Manning. 2020.
\newblock \href {https://openreview.net/forum?id=r1xMH1BtvB} {Electra:
  Pre-training text encoders as discriminators rather than generators}.
\newblock In \emph{International Conference on Learning Representations}.

\bibitem[{Dong et~al.(2019)Dong, Yang, Wang, Wei, Liu, Wang, Gao, Zhou, and
  Hon}]{dong2019unified}
Li~Dong, Nan Yang, Wenhui Wang, Furu Wei, Xiaodong Liu, Yu~Wang, Jianfeng Gao,
  Ming Zhou, and Hsiao-Wuen Hon. 2019.
\newblock \href
  {https://proceedings.neurips.cc/paper/2019/file/c20bb2d9a50d5ac1f713f8b34d9aac5a-Paper.pdf}
  {Unified language model pre-training for natural language understanding and
  generation}.
\newblock In \emph{Advances in Neural Information Processing Systems},
  volume~32. Curran Associates, Inc.

\bibitem[{Du and Cardie(2017)}]{du2017identifying}
Xinya Du and Claire Cardie. 2017.
\newblock \href {https://doi.org/10.18653/v1/D17-1219} {Identifying where to
  focus in reading comprehension for neural question generation}.
\newblock In \emph{Proceedings of the 2017 Conference on Empirical Methods in
  Natural Language Processing}, pages 2067--2073, Copenhagen, Denmark.
  Association for Computational Linguistics.

\bibitem[{Du et~al.(2017)Du, Shao, and Cardie}]{du2017learning}
Xinya Du, Junru Shao, and Claire Cardie. 2017.
\newblock \href {https://doi.org/10.18653/v1/P17-1123} {Learning to ask: Neural
  question generation for reading comprehension}.
\newblock In \emph{Proceedings of the 55th Annual Meeting of the Association
  for Computational Linguistics (Volume 1: Long Papers)}, pages 1342--1352,
  Vancouver, Canada. Association for Computational Linguistics.

\bibitem[{Duan et~al.(2017)Duan, Tang, Chen, and Zhou}]{duan2017question}
Nan Duan, Duyu Tang, Peng Chen, and Ming Zhou. 2017.
\newblock \href {https://doi.org/10.18653/v1/D17-1090} {Question generation for
  question answering}.
\newblock In \emph{Proceedings of the 2017 Conference on Empirical Methods in
  Natural Language Processing}, pages 866--874, Copenhagen, Denmark.
  Association for Computational Linguistics.

\bibitem[{Fan et~al.(2018)Fan, Wei, Li, Lan, and Huang}]{fan2018question-type}
Zhihao Fan, Zhongyu Wei, Piji Li, Yanyan Lan, and Xuanjing Huang. 2018.
\newblock \href {https://doi.org/10.24963/ijcai.2018/563} {A question type
  driven framework to diversify visual question generation}.
\newblock In \emph{Proceedings of the Twenty-Seventh International Joint
  Conference on Artificial Intelligence, {IJCAI-18}}, pages 4048--4054.
  International Joint Conferences on Artificial Intelligence Organization.

\bibitem[{Graves and Schmidhuber(2005)}]{graves2005bilstm}
Alex Graves and Jürgen Schmidhuber. 2005.
\newblock \href {https://doi.org/https://doi.org/10.1016/j.neunet.2005.06.042}
  {Framewise phoneme classification with bidirectional lstm and other neural
  network architectures}.
\newblock \emph{Neural Networks}, 18(5):602--610.
\newblock IJCNN 2005.

\bibitem[{Heilman and Smith(2010)}]{heilman2010good}
Michael Heilman and Noah~A. Smith. 2010.
\newblock \href {https://aclanthology.org/N10-1086} {Good question! statistical
  ranking for question generation}.
\newblock In \emph{Human Language Technologies: The 2010 Annual Conference of
  the North {A}merican Chapter of the Association for Computational
  Linguistics}, pages 609--617, Los Angeles, California. Association for
  Computational Linguistics.

\bibitem[{Hendrycks and Gimpel(2016)}]{hendrycks2016gelu}
Dan Hendrycks and Kevin Gimpel. 2016.
\newblock \href {http://arxiv.org/abs/1606.08415} {Bridging nonlinearities and
  stochastic regularizers with gaussian error linear units}.
\newblock \emph{ArXiv}, abs/1606.08415.

\bibitem[{Hochreiter and Schmidhuber(1997)}]{hochreiter1997lstm}
Sepp Hochreiter and J\"{u}rgen Schmidhuber. 1997.
\newblock \href {https://doi.org/10.1162/neco.1997.9.8.1735} {Long short-term
  memory}.
\newblock \emph{Neural Comput.}, 9(8):1735–1780.

\bibitem[{Holtzman et~al.(2020)Holtzman, Buys, Du, Forbes, and
  Choi}]{Holtzman2020The}
Ari Holtzman, Jan Buys, Li~Du, Maxwell Forbes, and Yejin Choi. 2020.
\newblock \href {https://openreview.net/forum?id=rygGQyrFvH} {The curious case
  of neural text degeneration}.
\newblock In \emph{International Conference on Learning Representations}.

\bibitem[{Hu et~al.(2018)Hu, Liu, Ma, Zhao, and Yan}]{hu2018aspect}
Wenpeng Hu, Bing Liu, Jinwen Ma, Dongyan Zhao, and Rui Yan. 2018.
\newblock Aspect-based question generation.
\newblock In \emph{International Conference of Representation Learning
  Workshop}.

\bibitem[{Kuhn(1955)}]{kuhn1955hungarian}
H.~W. Kuhn. 1955.
\newblock \href {https://doi.org/https://doi.org/10.1002/nav.3800020109} {The
  hungarian method for the assignment problem}.
\newblock \emph{Naval Research Logistics Quarterly}, 2(1-2):83--97.

\bibitem[{Kusner et~al.(2015)Kusner, Sun, Kolkin, and
  Weinberger}]{kusner2015word}
Matt~J. Kusner, Yu~Sun, Nicholas~I. Kolkin, and Kilian~Q. Weinberger. 2015.
\newblock From word embeddings to document distances.
\newblock In \emph{Proceedings of the 32nd International Conference on
  International Conference on Machine Learning - Volume 37}, ICML'15, page
  957–966. JMLR.org.

\bibitem[{Laban et~al.(2020)Laban, Canny, and Hearst}]{laban2020whats}
Philippe Laban, John Canny, and Marti~A. Hearst. 2020.
\newblock \href {https://doi.org/10.18653/v1/2020.acl-demos.43} {What{'}s the
  latest? a question-driven news chatbot}.
\newblock In \emph{Proceedings of the 58th Annual Meeting of the Association
  for Computational Linguistics: System Demonstrations}, pages 380--387,
  Online. Association for Computational Linguistics.

\bibitem[{Lin(2004)}]{lin-2004-rouge}
Chin-Yew Lin. 2004.
\newblock \href {https://aclanthology.org/W04-1013} {{ROUGE}: A package for
  automatic evaluation of summaries}.
\newblock In \emph{Text Summarization Branches Out}, pages 74--81, Barcelona,
  Spain. Association for Computational Linguistics.

\bibitem[{Liu et~al.(2020)Liu, Jiang, He, Chen, Liu, Gao, and Han}]{Liu2020On}
Liyuan Liu, Haoming Jiang, Pengcheng He, Weizhu Chen, Xiaodong Liu, Jianfeng
  Gao, and Jiawei Han. 2020.
\newblock \href {https://openreview.net/forum?id=rkgz2aEKDr} {On the variance
  of the adaptive learning rate and beyond}.
\newblock In \emph{International Conference on Learning Representations}.

\bibitem[{Lopez et~al.(2021)Lopez, Cruz, Cruz, and
  Cheng}]{lopez2021transformerqg}
Luis~Enrico Lopez, Diane~Kathryn Cruz, Jan Christian~Blaise Cruz, and Charibeth
  Cheng. 2021.
\newblock \href {https://arxiv.org/abs/2005.01107v4} {Transformer-based
  end-to-end question generation}.
\newblock \emph{In Proceedings of Pacific Rim International Conferences on
  Artificial Intelligence (PRICAI)}.

\bibitem[{Narayan et~al.(2020)Narayan, Sim{\~{o}}es, Ma, Craighead, and
  McDonald}]{narayan2020qurious}
Shashi Narayan, Gon{\c{c}}alo Sim{\~{o}}es, Ji~Ma, Hannah Craighead, and
  Ryan~T. McDonald. 2020.
\newblock \href {http://arxiv.org/abs/2004.11026} {{QURIOUS:} question
  generation pretraining for text generation}.
\newblock \emph{arXiv}, abs/2004.11026.

\bibitem[{Nema and Khapra(2018)}]{nema2018towards}
Preksha Nema and Mitesh~M. Khapra. 2018.
\newblock \href {https://doi.org/10.18653/v1/D18-1429} {Towards a better metric
  for evaluating question generation systems}.
\newblock In \emph{Proceedings of the 2018 Conference on Empirical Methods in
  Natural Language Processing}, pages 3950--3959, Brussels, Belgium.
  Association for Computational Linguistics.

\bibitem[{Papineni et~al.(2002)Papineni, Roukos, Ward, and
  Zhu}]{papineni2002bleu}
Kishore Papineni, Salim Roukos, Todd Ward, and Wei-Jing Zhu. 2002.
\newblock \href {https://doi.org/10.3115/1073083.1073135} {{B}leu: a method for
  automatic evaluation of machine translation}.
\newblock In \emph{Proceedings of the 40th Annual Meeting of the Association
  for Computational Linguistics}, pages 311--318, Philadelphia, Pennsylvania,
  USA. Association for Computational Linguistics.

\bibitem[{Puri et~al.(2020)Puri, Spring, Shoeybi, Patwary, and
  Catanzaro}]{puri2020training}
Raul Puri, Ryan Spring, Mohammad Shoeybi, Mostofa Patwary, and Bryan Catanzaro.
  2020.
\newblock \href {https://doi.org/10.18653/v1/2020.emnlp-main.468} {Training
  question answering models from synthetic data}.
\newblock In \emph{Proceedings of the 2020 Conference on Empirical Methods in
  Natural Language Processing (EMNLP)}, pages 5811--5826, Online. Association
  for Computational Linguistics.

\bibitem[{Qi et~al.(2020)Qi, Yan, Gong, Liu, Duan, Chen, Zhang, and
  Zhou}]{qi2020prophetnet}
Weizhen Qi, Yu~Yan, Yeyun Gong, Dayiheng Liu, Nan Duan, Jiusheng Chen, Ruofei
  Zhang, and Ming Zhou. 2020.
\newblock \href {https://doi.org/10.18653/v1/2020.findings-emnlp.217}
  {{P}rophet{N}et: Predicting future n-gram for
  sequence-to-{S}equence{P}re-training}.
\newblock In \emph{Findings of the Association for Computational Linguistics:
  EMNLP 2020}, pages 2401--2410, Online. Association for Computational
  Linguistics.

\bibitem[{Raffel et~al.(2020)Raffel, Shazeer, Roberts, Lee, Narang, Matena,
  Zhou, Li, and Liu}]{raffel2020exploring}
Colin Raffel, Noam Shazeer, Adam Roberts, Katherine Lee, Sharan Narang, Michael
  Matena, Yanqi Zhou, Wei Li, and Peter~J. Liu. 2020.
\newblock \href {http://jmlr.org/papers/v21/20-074.html} {Exploring the limits
  of transfer learning with a unified text-to-text transformer}.
\newblock \emph{Journal of Machine Learning Research}, 21(140):1--67.

\bibitem[{Rajpurkar et~al.(2018)Rajpurkar, Jia, and Liang}]{rajpurkar2018know}
Pranav Rajpurkar, Robin Jia, and Percy Liang. 2018.
\newblock \href {https://doi.org/10.18653/v1/P18-2124} {Know what you don{'}t
  know: Unanswerable questions for {SQ}u{AD}}.
\newblock In \emph{Proceedings of the 56th Annual Meeting of the Association
  for Computational Linguistics (Volume 2: Short Papers)}, pages 784--789,
  Melbourne, Australia. Association for Computational Linguistics.

\bibitem[{Rajpurkar et~al.(2016)Rajpurkar, Zhang, Lopyrev, and
  Liang}]{rajpurkar2016squad}
Pranav Rajpurkar, Jian Zhang, Konstantin Lopyrev, and Percy Liang. 2016.
\newblock \href {https://doi.org/10.18653/v1/D16-1264} {{SQ}u{AD}: 100,000+
  questions for machine comprehension of text}.
\newblock In \emph{Proceedings of the 2016 Conference on Empirical Methods in
  Natural Language Processing}, pages 2383--2392, Austin, Texas. Association
  for Computational Linguistics.

\bibitem[{Rubner et~al.(1998)Rubner, Tomasi, and Guibas}]{rubner1998metric}
Y.~Rubner, C.~Tomasi, and L.J. Guibas. 1998.
\newblock \href {https://doi.org/10.1109/ICCV.1998.710701} {A metric for
  distributions with applications to image databases}.
\newblock In \emph{Sixth International Conference on Computer Vision (IEEE Cat.
  No.98CH36271)}, pages 59--66.

\bibitem[{Rus and Lintean(2012)}]{rus2012comparison}
Vasile Rus and Mihai Lintean. 2012.
\newblock \href {https://aclanthology.org/W12-2018} {A comparison of greedy and
  optimal assessment of natural language student input using word-to-word
  similarity metrics}.
\newblock In \emph{Proceedings of the Seventh Workshop on Building Educational
  Applications Using {NLP}}, pages 157--162, Montr{\'e}al, Canada. Association
  for Computational Linguistics.

\bibitem[{Schlichtkrull and Cheng(2020)}]{sejr2020evaluating}
Michael~Sejr Schlichtkrull and Weiwei Cheng. 2020.
\newblock \href {http://arxiv.org/abs/2008.07291} {Evaluating for diversity in
  question generation over text}.
\newblock \emph{ArXiv}, abs/2008.07291.

\bibitem[{Scialom et~al.(2019)Scialom, Piwowarski, and
  Staiano}]{scialom2019self}
Thomas Scialom, Benjamin Piwowarski, and Jacopo Staiano. 2019.
\newblock \href {https://doi.org/10.18653/v1/P19-1604} {Self-attention
  architectures for answer-agnostic neural question generation}.
\newblock In \emph{Proceedings of the 57th Annual Meeting of the Association
  for Computational Linguistics}, pages 6027--6032, Florence, Italy.
  Association for Computational Linguistics.

\bibitem[{Shakeri et~al.(2020)Shakeri, Nogueira~dos Santos, Zhu, Ng, Nan, Wang,
  Nallapati, and Xiang}]{shakeri2020end}
Siamak Shakeri, Cicero Nogueira~dos Santos, Henghui Zhu, Patrick Ng, Feng Nan,
  Zhiguo Wang, Ramesh Nallapati, and Bing Xiang. 2020.
\newblock \href {https://doi.org/10.18653/v1/2020.emnlp-main.439} {End-to-end
  synthetic data generation for domain adaptation of question answering
  systems}.
\newblock In \emph{Proceedings of the 2020 Conference on Empirical Methods in
  Natural Language Processing (EMNLP)}, pages 5445--5460, Online. Association
  for Computational Linguistics.

\bibitem[{Sharma et~al.(2017)Sharma, El~Asri, Schulz, and
  Zumer}]{sharma2017nlgeval}
Shikhar Sharma, Layla El~Asri, Hannes Schulz, and Jeremie Zumer. 2017.
\newblock \href {http://arxiv.org/abs/1706.09799} {Relevance of unsupervised
  metrics in task-oriented dialogue for evaluating natural language
  generation}.
\newblock \emph{CoRR}, abs/1706.09799.

\bibitem[{Song et~al.(2017)Song, Wang, and Hamza}]{song2017unified}
Linfeng Song, Zhiguo Wang, and Wael Hamza. 2017.
\newblock A unified query-based generative model for question generation and
  question answering.
\newblock \emph{ArXiv}, abs/1709.01058.

\bibitem[{Subramanian et~al.(2018)Subramanian, Wang, Yuan, Zhang, Trischler,
  and Bengio}]{subramanian2018neural}
Sandeep Subramanian, Tong Wang, Xingdi Yuan, Saizheng Zhang, Adam Trischler,
  and Yoshua Bengio. 2018.
\newblock \href {https://doi.org/10.18653/v1/W18-2609} {Neural models for key
  phrase extraction and question generation}.
\newblock In \emph{Proceedings of the Workshop on Machine Reading for Question
  Answering}, pages 78--88, Melbourne, Australia. Association for Computational
  Linguistics.

\bibitem[{Sun et~al.(2018)Sun, Liu, Lyu, He, Ma, and Wang}]{sun2018answer}
Xingwu Sun, Jing Liu, Yajuan Lyu, Wei He, Yanjun Ma, and Shi Wang. 2018.
\newblock \href {https://doi.org/10.18653/v1/D18-1427} {Answer-focused and
  position-aware neural question generation}.
\newblock In \emph{Proceedings of the 2018 Conference on Empirical Methods in
  Natural Language Processing}, pages 3930--3939, Brussels, Belgium.
  Association for Computational Linguistics.

\bibitem[{Tang et~al.(2017)Tang, Duan, Qin, and Zhou}]{duyu2017question}
Duyu Tang, Nan Duan, Tao Qin, and Ming Zhou. 2017.
\newblock Question answering and question generation as dual tasks.
\newblock \emph{ArXiv}, abs/1706.02027.

\bibitem[{Varanasi et~al.(2020)Varanasi, Amin, and
  Neumann}]{varanasi2020copybert}
Stalin Varanasi, Saadullah Amin, and Guenter Neumann. 2020.
\newblock \href {https://doi.org/10.18653/v1/2020.nlp4convai-1.3}
  {{C}opy{BERT}: A unified approach to question generation with
  self-attention}.
\newblock In \emph{Proceedings of the 2nd Workshop on Natural Language
  Processing for Conversational AI}, pages 25--31, Online. Association for
  Computational Linguistics.

\bibitem[{Wang et~al.(2020)Wang, Cho, and Lewis}]{wang2020asking}
Alex Wang, Kyunghyun Cho, and Mike Lewis. 2020.
\newblock \href {https://doi.org/10.18653/v1/2020.acl-main.450} {Asking and
  answering questions to evaluate the factual consistency of summaries}.
\newblock In \emph{Proceedings of the 58th Annual Meeting of the Association
  for Computational Linguistics}, pages 5008--5020, Online. Association for
  Computational Linguistics.

\bibitem[{Wang et~al.(2019)Wang, Wei, Fan, Liu, and
  Huang}]{wang2019multi-agent}
Siyuan Wang, Zhongyu Wei, Zhihao Fan, Yang Liu, and Xuanjing Huang. 2019.
\newblock \href {https://doi.org/10.1609/aaai.v33i01.33017168} {A multi-agent
  communication framework for question-worthy phrase extraction and question
  generation}.
\newblock \emph{Proceedings of the AAAI Conference on Artificial Intelligence},
  33(01):7168--7175.

\bibitem[{Wang et~al.(2017)Wang, Yuan, and Trischler}]{wang2017joint}
Tong Wang, Xingdi Yuan, and Adam Trischler. 2017.
\newblock A joint model for question answering and question generation.
\newblock \emph{ArXiv}, abs/1706.01450.

\bibitem[{Wang et~al.(2018)Wang, Liu, Huang, and Nie}]{wang2018learning}
Yansen Wang, Chenyi Liu, Minlie Huang, and Liqiang Nie. 2018.
\newblock \href {https://doi.org/10.18653/v1/P18-1204} {Learning to ask
  questions in open-domain conversational systems with typed decoders}.
\newblock In \emph{Proceedings of the 56th Annual Meeting of the Association
  for Computational Linguistics (Volume 1: Long Papers)}, pages 2193--2203,
  Melbourne, Australia. Association for Computational Linguistics.

\bibitem[{Willis et~al.(2019)Willis, Davis, Ruan, Manoharan, Landay, and
  Brunskill}]{willis2019keyphrase}
Angelica Willis, Glenn Davis, Sherry Ruan, Lakshmi Manoharan, James Landay, and
  Emma Brunskill. 2019.
\newblock \href {https://doi.org/10.1145/3330430.3333636} {Key phrase
  extraction for generating educational question-answer pairs}.
\newblock In \emph{Proceedings of the Sixth (2019) ACM Conference on Learning @
  Scale}, L@S '19, New York, NY, USA. Association for Computing Machinery.

\bibitem[{Wu et~al.(2020)Wu, Jiang, and Wu}]{wu2020question}
Xiuyu Wu, Nan Jiang, and Yunfang Wu. 2020.
\newblock \href {https://doi.org/10.18653/v1/2020.ngt-1.8} {A question type
  driven and copy loss enhanced framework for answer-agnostic neural question
  generation}.
\newblock In \emph{Proceedings of the Fourth Workshop on Neural Generation and
  Translation}, pages 69--78, Online. Association for Computational
  Linguistics.

\bibitem[{Yuan et~al.(2017)Yuan, Wang, Gulcehre, Sordoni, Bachman, Zhang,
  Subramanian, and Trischler}]{yuan2017machine}
Xingdi Yuan, Tong Wang, Caglar Gulcehre, Alessandro Sordoni, Philip Bachman,
  Saizheng Zhang, Sandeep Subramanian, and Adam Trischler. 2017.
\newblock \href {https://doi.org/10.18653/v1/W17-2603} {Machine comprehension
  by text-to-text neural question generation}.
\newblock In \emph{Proceedings of the 2nd Workshop on Representation Learning
  for {NLP}}, pages 15--25, Vancouver, Canada. Association for Computational
  Linguistics.

\bibitem[{Yuan et~al.(2020)Yuan, Wang, Meng, Thaker, Brusilovsky, He, and
  Trischler}]{yuan-etal-2020-one}
Xingdi Yuan, Tong Wang, Rui Meng, Khushboo Thaker, Peter Brusilovsky, Daqing
  He, and Adam Trischler. 2020.
\newblock \href {https://doi.org/10.18653/v1/2020.acl-main.710} {One size does
  not fit all: Generating and evaluating variable number of keyphrases}.
\newblock In \emph{Proceedings of the 58th Annual Meeting of the Association
  for Computational Linguistics}, pages 7961--7975, Online. Association for
  Computational Linguistics.

\bibitem[{Zhang* et~al.(2020)Zhang*, Kishore*, Wu*, Weinberger, and
  Artzi}]{Zhang2020BERTScore}
Tianyi Zhang*, Varsha Kishore*, Felix Wu*, Kilian~Q. Weinberger, and Yoav
  Artzi. 2020.
\newblock \href {https://openreview.net/forum?id=SkeHuCVFDr} {Bertscore:
  Evaluating text generation with bert}.
\newblock In \emph{International Conference on Learning Representations}.

\bibitem[{Zhao et~al.(2019)Zhao, Peyrard, Liu, Gao, Meyer, and
  Eger}]{zhao2019moverscore}
Wei Zhao, Maxime Peyrard, Fei Liu, Yang Gao, Christian~M. Meyer, and Steffen
  Eger. 2019.
\newblock \href {https://doi.org/10.18653/v1/D19-1053} {{M}over{S}core: Text
  generation evaluating with contextualized embeddings and earth mover
  distance}.
\newblock In \emph{Proceedings of the 2019 Conference on Empirical Methods in
  Natural Language Processing and the 9th International Joint Conference on
  Natural Language Processing (EMNLP-IJCNLP)}, pages 563--578, Hong Kong,
  China. Association for Computational Linguistics.

\bibitem[{Zhao et~al.(2018)Zhao, Ni, Ding, and Ke}]{zhao2018paragraph}
Yao Zhao, Xiaochuan Ni, Yuanyuan Ding, and Qifa Ke. 2018.
\newblock \href {https://doi.org/10.18653/v1/D18-1424} {Paragraph-level neural
  question generation with maxout pointer and gated self-attention networks}.
\newblock In \emph{Proceedings of the 2018 Conference on Empirical Methods in
  Natural Language Processing}, pages 3901--3910, Brussels, Belgium.
  Association for Computational Linguistics.

\bibitem[{Zhou et~al.(2017)Zhou, Yang, Wei, Tan, Bao, and
  Zhou}]{zhou2017neural}
Qingyu Zhou, Nan Yang, Furu Wei, Chuanqi Tan, Hangbo Bao, and Ming Zhou. 2017.
\newblock Neural question generation from text: {A} preliminary study.
\newblock \emph{In National CCF Conference on Natural Language Processing and
  Chinese Computing}.

\bibitem[{Zhou et~al.(2019)Zhou, Zhang, and Wu}]{zhou2019questiontype}
Wenjie Zhou, Minghua Zhang, and Yunfang Wu. 2019.
\newblock \href {https://doi.org/10.18653/v1/D19-1622} {Question-type driven
  question generation}.
\newblock In \emph{Proceedings of the 2019 Conference on Empirical Methods in
  Natural Language Processing and the 9th International Joint Conference on
  Natural Language Processing (EMNLP-IJCNLP)}, pages 6032--6037, Hong Kong,
  China. Association for Computational Linguistics.

\bibitem[{Zhu et~al.(2018)Zhu, Lu, Zheng, Guo, Zhang, Wang, and
  Yu}]{zhu2018texygen}
Yaoming Zhu, Sidi Lu, Lei Zheng, Jiaxian Guo, Weinan Zhang, Jun Wang, and Yong
  Yu. 2018.
\newblock \href {https://doi.org/10.1145/3209978.3210080} {Texygen: A
  benchmarking platform for text generation models}.
\newblock In \emph{The 41st International ACM SIGIR Conference on Research;
  Development in Information Retrieval}, SIGIR '18, page 1097–1100, New York,
  NY, USA. Association for Computing Machinery.

\end{thebibliography}
